\documentclass[10pt]{article}
\usepackage{cite} 
\usepackage{url}  
\usepackage{multicol} 
\usepackage[utf8]{inputenc}
\usepackage{graphicx}
\usepackage{tabularx}
\usepackage{amsfonts}
\usepackage{amsmath}
\usepackage{amssymb}
\usepackage{graphicx}
\usepackage{multirow}
\usepackage{varwidth}
\usepackage[table,xcdraw]{xcolor}

\newcommand{\eqdef}{\overset{\mathrm{def}}{\joinrel=}}

\begin{document}

\begingroup  
\centering
\LARGE An Analytic Framework for \\ Maritime Situation Analysis\\[1em]
\large Hamed Yaghoubi Shahir, Uwe Gl\"asser, Amir Yaghoubi Shahir\\[0.2em]
\small Software Technology Lab, Simon Fraser University, Burnaby, BC, Canada\\[0.5em]
\large Hans Wehn\\[0.2em]
\small MDA Corporation, Richmond, BC, Canada\\[2em]
\endgroup

\begin{abstract}
Maritime domain awareness is critical for protecting sea lanes, ports, harbors, offshore structures and critical infrastructures against common threats and illegal activities. Limited surveillance resources constrain maritime domain awareness and compromise full security coverage at all times. This situation calls for innovative intelligent systems for interactive situation analysis to assist marine authorities and security personal in their routine surveillance operations. In this article, we propose a novel situation analysis framework to analyze marine traffic data and differentiate various scenarios of vessel engagement for the purpose of detecting anomalies of interest for marine vessels that operate over some period of time in relative proximity to each other. The proposed framework views vessel behavior as probabilistic processes and uses machine learning to model common vessel interaction patterns. We represent patterns of interest as left-to-right Hidden Markov Models and classify such patterns using Support Vector Machines.
\end{abstract}
\section{Introduction}

We consider marine traffic scenarios where two or more boats or ships engage in some form of {coordinated interaction} while operating over some period of time in relative proximity to each other.
Maritime surveillance data is interpreted as sequences of discrete observations (e.g., position, heading, speed) over time that render vessel {\it trajectories}. We model, analyze and differentiate vessel interaction scenarios based on a combination of kinematic, geospatial and other features for the purpose of detecting anomalies of interest in near real time. 
We generalize here the classical concept of {\it rendezvous} to encompass any number and type of marine vessels, including all ships and boats regardless of their size and function, as well as offshore structures like oil and gas rigs, and to any kind of interaction between vessels that operate in relative proximity to each other or an offshore structure over some period of time.
Beyond kinematic features (e.g., speed), we also analyze geospatial aspects (e.g., operating in restricted areas), geometric aspects (e.g., proximity to other vessels), contextual information (e.g., vessel size and cargo type) and domain knowledge (e.g., possible impact of a scenario) to provide a more holistic picture of a situation as it unfolds. 

\section{The Proposed Framework}
\label{classification-and-detection-framework}
The classification and detection process is divided into three  separate yet {logically  connected phases: Engagement Detection, Scenario Detection, and Anomaly Detection (see Figure~\ref{fig:approach}).

\begin{figure*}[h]
   \centering
   \includegraphics[width=0.95\textwidth]{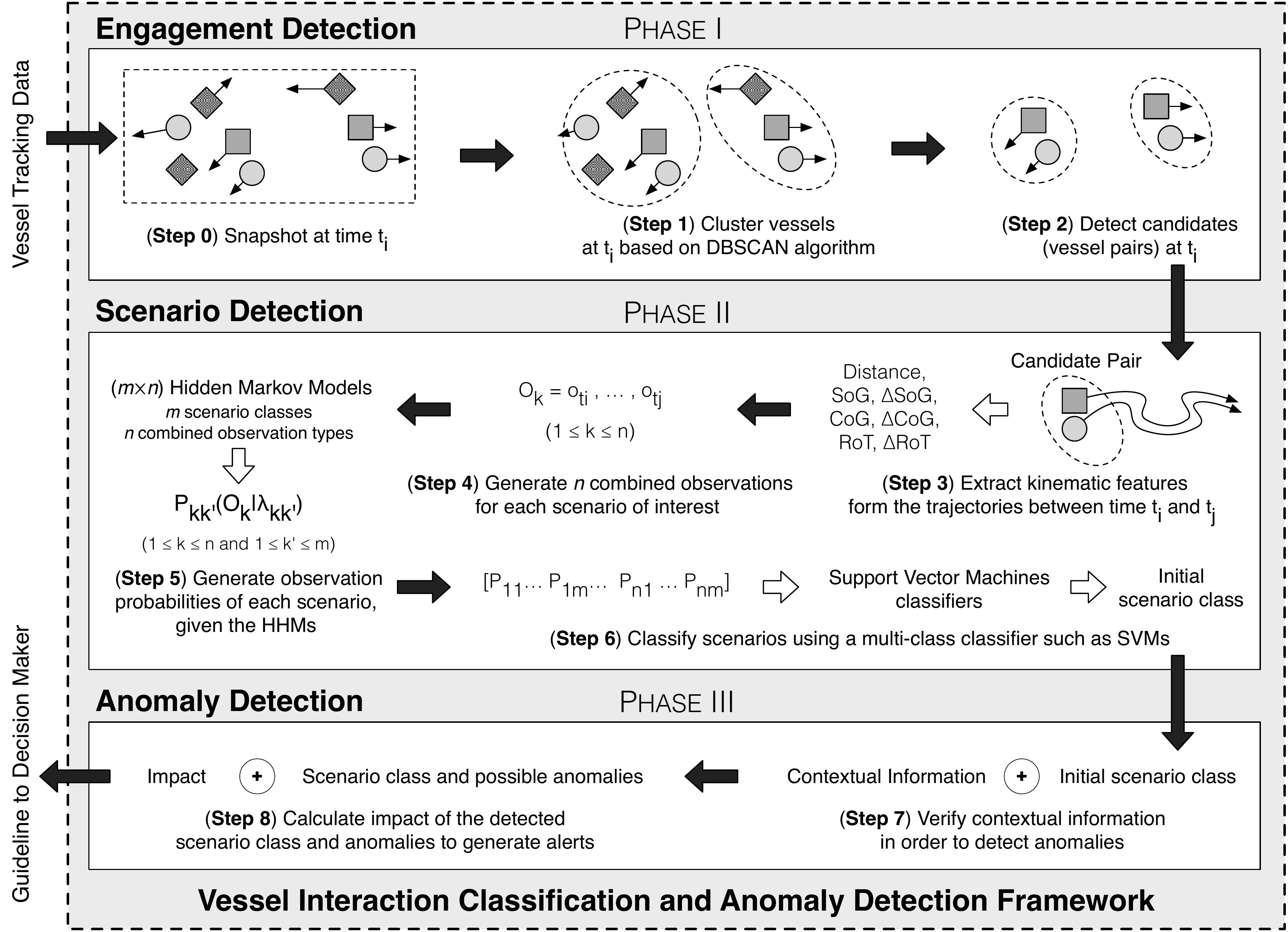} 
   \caption{Overview of the Proposed Framework}
   \label{fig:approach}
\end{figure*}

\subsection{Engagement Detection}
\label{sec:engagement-detection}
The purpose of this phase, which is the first of the three phases, is to identify vessels that are within a certain proximity range to each other and determine with a high confidence whether they are effectively {\it engaging} in any interaction scenario. The challenge is to monitor this situation in real time on a continuous basis for hundreds, even thousands, of moving marine vessels.
However, many vessels operate in some distance to each other and are unlikely able to get close enough within a foreseeable time period of a few minutes. Thus, a proximity sensitive approach, limiting observable interactions to a proximity range $\delta$, is promising. \\

\noindent {\bf {Step~1) Cluster Vessels}:} 
The first step of this phase identifies vessels that are within a close enough distance to each other. 
For clustering we use the popular DBSCAN \cite{DBscan} algorithm as a density-based clustering approach with parameters {\it epsilon = $\delta$} and {\it minPoints = 2}. \\

\noindent {\bf  {Step~2) Detect Candidates}:}
The second step evaluates for all pairs of vessels belonging to the same {\it cluster} conditions indicating engagement based on kinematic features. We have identified a number of conditions that indicate engagement of two vessels in an interaction scenario in addition to the one exemplified in the following:

\smallskip
\noindent${\sf Engaging}(v_i, v_j: \mbox{\it Vessel}) \eqdef$\\
\noindent $~~~{\sf SoG}(v_i) < \theta_{v_i} \land\; {\sf SoG}(v_j) < \theta_{v_j} \land\; ({\sf Proximity}(v_i, v_j, \delta') \lor \; {\sf Converging}(v_i, v_j, \delta', \tau))$\\

\noindent The outcome of the Engagement Detection phase is a set of {\it candidates}, pairs of vessels which are close enough and possibly engaging in an interaction scenario.

\subsection{Scenario Detection}
\label{sec:scenario-detection}
In the Scenario Detection phase, the outcome of the Engagement Detection, phase is used to analyze candidate trajectories. Hidden Markov Models (HMMs) \cite{Rabiner1989}, along with Support Vector Machines (SVMs) \cite{Cortes1995} as multi-class classifiers, are used to classify and detect, with notable likelihood, scenarios of interest.\\

\noindent {\bf  {Step 3) Extract Kinematic Features}:}
Observable entities of this step are candidate pairs of vessels for which their combined behavior is to be analyzed. This step extracts the movement features of candidates which are either atomic or composite. An atomic feature refers to a single characteristic of an individual vessel $v$ in its corresponding trajectory at a certain time $t$, such as {\it SoG(v, t)}; while a composite feature refers to the composition of two identical atomic features of two different vessels $v_i$ and $v_i$ at time $t$, such as $\Delta SoG(v_i, v_j, t) = SoG(v_i, t) - SoG(v_j, t)$ or $Distance(v_i, v_j, t)$. Table \ref{tbl:observed-variables} lists the atomic and composite features used here.

 \begin{table}[h]
 \centering
 \caption{Atomic and Composite Kinematic Features}
\footnotesize        
 \begin{tabular}{c|c|l|}
\cline{2-3}
 & \multicolumn{1}{c|}{{Feature}} & \multicolumn{1}{c|}{{Description}} \\  \hline
\multicolumn{1}{|l|}{{\multirow{3}{*}{{Atomic}}}} & $SoG$ & speed over ground of a vessel \\  \cline{2-3}
\multicolumn{1}{|l|}{}  & $CoG$ & course over ground of a vessel\\ \cline{2-3}
\multicolumn{1}{|l|}{}  & $RoT$ & rate of turn of a vessel\\ 
\hline
\multicolumn{1}{|l|}{  {\multirow{4}{*}{{Composite}}}} & $Distance$ & distance between two vessels \\ \cline{2-3}
\multicolumn{1}{|l|}{}  & $\Delta SoG$ & difference in SoG of two vessels\\  \cline{2-3}
\multicolumn{1}{|l|}{} & $\Delta CoG$ & difference in CoG of two vessels\\ \cline{2-3}
\multicolumn{1}{|l|}{} & $\Delta RoT$ & difference in RoT of two vessels\\ 
\hline
 \end{tabular}
 \label{tbl:observed-variables}
 \end{table}

\noindent {\bf {Step 4) Generate Combined Observations}:}
This step generates a collection of $n$ {\it combined observations} for each candidate. Combined observation of two vessels $v_i$ and $v_j$, $O(v_i, v_j, t, t')$, refers to a sequence of {\it observation points} (a number of atomic or composite {\it features} of vessel $v_i$ and $v_j$ at a certain time $t$ such as $[\Delta SoG(v_i, v_j, t), SoG(v_i, t)]$) for the same set of features $o(v_i, v_j, t), \dots, o(v_i, v_j, t')$ between time $t$ and $t'$.\\ 

\noindent {\bf {Step 5) Generate Observation Probabilities using Markov Models}:}
{Given a number of different combined observations $O_x$ and HMMs $\lambda_y$, we compute $P(O_x | \lambda_y)$, the probabilities of the combined observations, given the models. The number of combined observation types and HMMs to be used for the scenario classification depends on the classification strategy.} \\

\noindent {\bf {Step 6) Classify Scenario}:}
Each of the $m$ scenario classes is modeled by $n$ HMMs, each trained based on one of the $n$ different types of {\it combined observations}. This results in a set of $n \times m$ HMMs, $\lambda_{1,1}, \dots, \lambda_{1,n}, \dots, \lambda_{m,1}, \dots, \lambda_{m,n}$. Multi-class SVMs are used to classify scenarios based on these $n \times m$ probability values. \\

\noindent It is noteworthy that this phase determines the scenario class solely on the basis of four kinematic features and their combinations. Detecting both, normal and anomalous scenarios, can significantly reduce the overall workload of human operators. Although using kinematic features is necessary, this alone is not sufficient to provide high confidence feedback.

\subsection{Anomaly Detection}
\label{sec:anomaly-detection}

In the third phase, the outcome of Scenario Detection is linked to {\it contextual information} and {\it domain knowledge}. This extends basic scenario detection beyond ``what you see'' by observing kinematic features to also take into account ``what you know''. 
Intuitively, this phase is highly interactive as domain expert knowledge plays a crucial role and is invaluable for `connecting the dots'.\\

\noindent {\bf Step 7) Verify Contextual Information}:}
{One can define basic domain knowledge and contextual information of a scenario in terms of first-order logic clauses.}  
Once Scenario Detection has identified 
the scenario class, the contextual information and background knowledge 
related to the scenario are considered for the reasoning process (e.g., resolution, answer set programming, etc.) to verify and possibly revise the class of the observed scenario.\\

\noindent {\bf {Step 8) Calculate Impact}:}
Domain experts define estimated {\it impact} values ($0 \le \mbox{\it impact} \le 1$) for each scenario class which is providing a measure for the estimated damage associated with a scenario. \\

\noindent Any {\it conflict} between the outcomes of the Scenario Detection and Anomaly Detection phases may be a possible anomaly and needs to be reported. Ultimately, the outcome of the {third and final} phase is decision-making guidelines and alerts for human operators based on the {\it impact} of the situation. 

\section{Experimental Evaluation}
We base our experimental evaluation on AIS data collected by U.S.\ Coast Guard\footnote{http://www.marinecadastre.gov} using onboard navigation safety devices that transmit vessel position, speed, course, etc. The UTM zones 1 to 11 for all year 2009 datasets which cover almost the entire west coast of North America are used for the experimental analysis. We consider here five classes of scenarios and Table \ref{tbl:occurrence} shows the number of occurrences of each class extracted from the AIS dataset.

\indent{\it Class-A)} a cargo/tanker vessel and a towing/tug vessel;\\
\indent{\it Class-B)} a cargo/tanker/passenger vessel and a pilot vessel;\\
\indent{\it Class-C)} two tanker vessels;\\
\indent{\it Class-D)} two passenger vessels;\\
\indent{\it Class-E)} two search and rescue vessels.\\[-18pt]

\begin{table}[h]
\footnotesize
\caption{Number of Occurrences Extracted for Each Scenario Class}
\begin{center}
\begin{tabular}{|c|c|c|c|c|c|}
\hline
Scenario Class & A     & B    & C   & D   & E   \\ \hline
\#Occurrence  & 44188 & 2372 & 252 & 671 & 378 \\ \hline
\end{tabular}
\end{center}
\label{tbl:occurrence}
\end{table}

\subsection{Experimental Design \& Performance Analysis}

The samples of each scenario class are partitioned into 10 non-overlapping sets and {run 10-fold cross validation by considering 90\% of the samples for training and 10\% for testing}. This way, it is guaranteed that the portions of scenario classes in all 10 training sets are (almost) equal; and, more importantly, there are two disjoint sets of sample scenarios for training and testing. 
We experimentally evaluate the effectiveness of using HMMs and SVMs for scenario classification. We use the open-source HMM Toolbox \cite{HMM-Toolbox} and the open-source library LIBSVM \cite{LibSVM} for training and testing.

\begin{table}[h]
\footnotesize
\caption{Confusion Matrix, and Precision and Recall of  Each Individual Class}
\begin{center}
\begin{tabular}{lc|c|c|c|c|c|}

\cline{3-7}
                                                   & \multicolumn{1}{l|}{}                           & \multicolumn{5}{c|}{Predicted Class}                                          \\ \cline{3-7} 
                                                   &                                                      & A             & B             & C             & D             & E             \\ \hline
\multicolumn{1}{|l|}{\multirow{5}{*}{\rotatebox[origin=c]{90}{True Class}}} & \multicolumn{1}{c|}{\multirow{0}{*}{A}}                           & 43592         & 473           & 31            & 60            & 32            \\ \cline{2-7} 
\multicolumn{1}{|l|}{}                             & \multicolumn{1}{c|}{\multirow{0}{*}{B}}                              & 688           & 1673          & 1             & 8             & 2             \\ \cline{2-7} 
\multicolumn{1}{|l|}{}                             & \multicolumn{1}{c|}{\multirow{0}{*}{C}}                            & 56            & 0             & 196           & 0             & 0             \\ \cline{2-7} 
\multicolumn{1}{|l|}{}                             & \multicolumn{1}{c|}{\multirow{0}{*}{D}}                             & 89            & 15            & 0             & 550           & 17            \\ \cline{2-7} 
\multicolumn{1}{|l|}{}                             & \multicolumn{1}{c|}{\multirow{0}{*}{E}}                           & 72            & 13            & 0             & 23            & 270           \\ \hline \\[-6pt]  \cline{2-7}
\multicolumn{1}{l|}{}                              & \multicolumn{1}{c|}{\multirow{0}{*}{{Precision}}}                    & 97.97\%       & { 76.95\%} & { 85.96\%} & { 85.80\%} & { 84.11\%} \\ \cline{2-7} 
\multicolumn{1}{l|}{}                              & \multicolumn{1}{c|}{\multirow{0}{*}{{Recall}}}                      & { 98.65\%} & 70.53\%       & 77.78\%       & 81.97\%       & 71.43\%       \\ \cline{2-7} 

\end{tabular}
\end{center}
\label{tbl:confusion-matrix}
\end{table}

\section{Conclusions}
\label{sec:conclusion}
Intelligent systems for interactive situation analysis and decision support in real-world situation assessment require innovative methodical approaches to economically develop robust and scalable solutions. We epitomize this idea here for the maritime domain, proposing a generalized model for analyzing vessel interaction patterns, where each pattern refers to a family of multi-vessel scenarios with common kinematic and typically also non-kinematic characteristics.
This analytic approach is novel and combines qualitative with quantitative modeling aspects into a 3-phase classification and detection framework serving two important practical purposes: {\it 1)} reducing the overall volume of observation data to be monitored and analyzed by coast guard services and marine authorities; and {\it 2)} providing a prioritized list of highly suspicious and critical scenarios ranked according to the anticipated impact, which then calls for human attention.
In fact, each of the three phases serves as a filter to reduce the volume to be processed by the subsequent phase, effectively performing a closer inspection of increasingly more relevant threats to safety and security.

In abstract terms, filtering is done for all vessels that engage in some form of observable interaction through: {\it 1)} fusing the kinematic features of these vessels into a uniformly analyzable entity; {\it 2)} fusing contextual information related to these vessels for a deeper analysis of the entity. The experimental results show that the accuracy of our proposed approach---using a collection of HMMs together with SVMs---is 96.7\%, making this framework promising for practical use in real-world situation analysis for maritime domain awareness. Although the experiments presented here focus on the west coast of North America, the expectation is that the same framework produces similar results for any coastal waters of North America and beyond. In our future work we will extend experiments to include the East Coast and the Golf of Mexico.

\end{document}